# Data-Prep-Kit: getting your data ready for LLM application development

David Wood[§], Boris Lublinsky[‡], Alexy Roytman[†], Shivdeep Singh[¦], Constantin Adam[§], Abdulhamid Adebayo[§], Sungeun An[§§], Yuan Chi Chang[§], Xuan-Hong Dang[§], Nirmit Desai[§], Michele Dolfi[*], Hajar Emami-Gohari[§], Revital Eres[†], Takuya Goto[¶], Dhiraj Joshi[§], Yan Koyfman[§], Mohammad Nassar[†], Hima Patel[¦], Paramesvaran Selvam[¦], Yousaf Shah[§], Saptha Surendran[¦], Daiki Tsuzuku[¶], Petros Zerfos[§], and Shahrokh Daijavad[§§]

emails: {dawood@us.ibm.com, blublinsk@ibm.com, roytman@il.ibm.com, shivdeep.singh@ibm.com, cmadam@us.ibm.com, hamid.adebayo@ibm.com, sungeun.an@ibm.com, yuanchi@us.ibm.com, xuan-hong.dang@ibm.com, nirmit.desai@us.ibm.com, dol@zurich.ibm.com, hajar.emami@ibm.com, eres@il.ibm.com, tkyg@jp.ibm.com, djoshi@us.ibm.com, koyfman@us.ibm.com, Mohammad.Nassar@ibm.com, himapatel@in.ibm.com, parselva@in.ibm.com, syshah@us.ibm.com, saptha.surendran@ibm.com, dtsuzuku@jp.ibm.com, pzerfos@us.ibm.com, shahrokh@us.ibm.com}

§ IBM Research, Yorktown Heights, New York, USA
§§ Almaden Research Center IBM, San Jose, CA, USA
‡ IBM Research, Mulhaddart, Ireland
† IBM Research Israel, Haifa, Israel
¦ IBM Research India, Bangalore, India
¶ IBM Software, Tokyo, Japan
* IBM Research, Zurich, Switzerland

*Abstract*— Data preparation is the first and a very important step towards any Large Language Model (LLM) development. This paper introduces an easy-to-use, extensible, and scale-flexible open-source data preparation toolkit called *Data Prep Kit* (*DPK*). *DPK* is architected and designed to enable users to scale their data preparation to their needs. With DPK they can prepare data on a local machine or effortlessly scale to run on a cluster with thousands of CPU Cores. DPK comes with a highly scalable, yet extensible set of modules that transform natural language and code data. If the user needs additional transforms, they can be easily developed using extensive DPK support for transform creation. These modules can be used independently or pipelined to perform a series of operations. In this paper, we describe DPK architecture and show its performance from a small scale to a very large number of CPUs. The modules from *DPK* have been used for the preparation of Granite Models [1] [2]. We believe DPK is a valuable contribution to the AI community to easily prepare data to enhance the performance of their LLM models or to fine-tune models with Retrieval-Augmented Generation (RAG).

## I. Introduction

Data is the starting point for building any Large Language Model (LLM) application. It is well understood that the quality of a model is heavily influenced by the quality of data [3] [4]. However, data preparation is still seen as one of the most difficult steps in the data and LLM lifecycle. This is primarily due to many factors, one of which is that the quality of the data is hard to assess upfront and hence the processing needs to be done to the data before using it for LLM purposes. Data quality is discovered as part of the data and model lifecycle. Most data scientists start with some common data preparation steps like data ingestion, extraction into a standardized format, and cleaning based on issues that they may be aware of in the data. However, often, these are not sufficient, and hidden data issues persist in the data. These hidden data issues impact the quality of the model and are typically discovered by analyzing the model results and tying them back to the data. This iterative data and model debugging is time-consuming, compute-intensive, cumbersome, and makes the whole development process slow. This problem is amplified by the scale of the data, thereby making manual inspections difficult or even impossible. Moreover, as new use cases emerge for building LLM applications, they bring in new data challenges that may be specific to each use case. Another challenge with industrial LLM projects is that the time and effort needed to take a project from successful proof of concept to product-ready deliverable is very high. The proof-of-concept stage is usually done with small datasets with a focus on validating the technical feasibility and return on investment. However, to take a project from proof of concept to production may require serious software engineering efforts, which may be outside the skills of a data scientist. All these challenges have been observed as part of our interactions with various teams building LLM applications in our organization. To overcome these challenges, we introduce a new open-source toolkit called data-prep-kit, which can be accessed at https://github.com/IBM/data-prep-kit. DPK is designed and architected to achieve the following goals:

1. The toolkit should provide support for using data preparation modules for various data modalities in a consistent fashion.

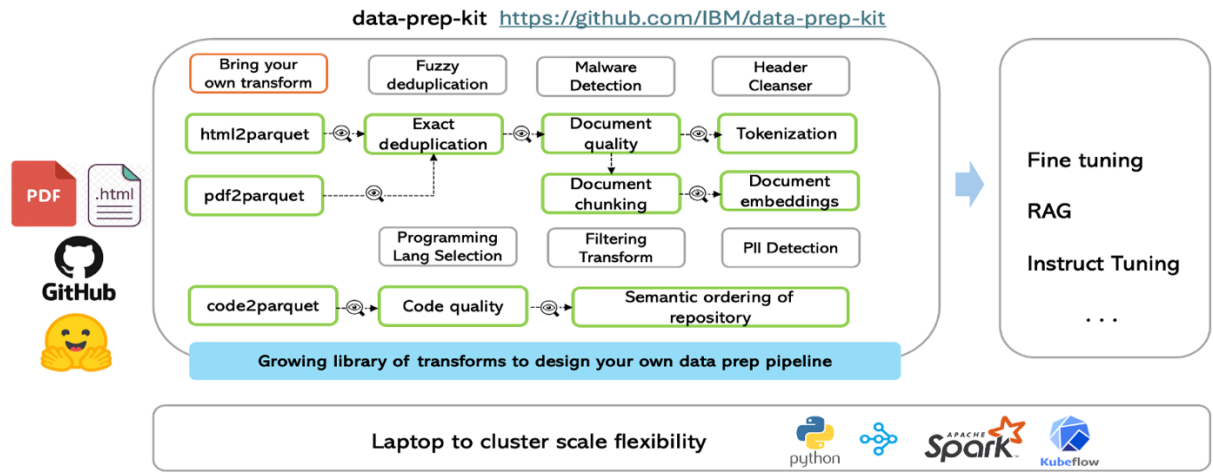

Fig. 1 Overview of data-prep-kit

2. The toolkit should support a user in both their proof-of-concept stage as well as the production stage. Thus, it should offer scale flexibility to run the data preparation modules from a laptop to a cluster.

3. The toolkit should support various personas like data scientists, data engineers, etc. These personas may have varying skills for building scalable data preparation pipelines, especially with frameworks like Ray [5] and Spark [6]. We thus form a goal for ourselves that the toolkit should be usable by anyone without needing deep knowledge of distributed computing, including Kubernetes, Ray, and Spark.

4. The toolkit should automate the capabilities so that one can run data preparation pipelines in a no-code fashion, by using Kubeflow Pipelines KFP [7] UI.

5. Finally, the framework should easily allow users to add a new data preparation module, without having extensive knowledge of Ray or Spark.

Fig. 1 shows a high-level overview of DPK. As can be seen in the figure, this toolkit offers many out-of-the-box data preparation modules (transforms) that can be connected to form data preparation pipelines serving a user's specific needs. For example, pipeline shown in Fig. 1 prepares data for a Retrieval Augmented Generation (RAG) task by combining the following transforms: *pdf2parquet*, *exact-deduplication*, *document-quality* check, *document-chunking*, and building *document-embeddings* that can go as part of a vector store. In Section IV, we will discuss the ready-to-use transforms currently provided by DPK. While having many out-of-the-box transforms is a big advantage, the main power of our toolkit lies in the support it provides for the user to create custom transforms, scale them from laptop to cluster by supporting various runtimes like Ray, Spark and automate transform execution leveraging KubeFlow Pipelines (KFP).

Our contribution in this paper via DPK can be summarized as:

1. Platform Flexibility: Support for flexible computing options from laptops to large Kubernetes clusters by leveraging Ray and Spark runtimes.

2. Extendibility: Novel data-processing-lib which is a framework that abstracts out the details of Ray and Spark, thereby allowing easy addition of new data prep modules with minimal Ray/Spark skills.

3. Automation with scalability: Automation via KubeFlow Pipelines that enables running data prep pipelines in a no-code mode through the UI, thereby making it usable across various personas.

After a review of closely related work in Section II, we discuss the design of our toolkit in Section III, followed by a list of available transforms in Section IV. Next, we discuss how we achieve automation via KubeFlow in Section V, followed by how a user can add a new transform in Section VI. In the experimental results Section VII, we show real execution results of using our toolkit for data preparation ranging from small-scale data sets to very large-scale data, going to terabytes. Our results for large-scale data analysis are derived from data preparation runs for training our Granite models [1].

## II. RELATED WORK

There are other open-source projects for data processing and preparation of LLM applications, e.g., BigCode [8], DataTrove [9], Dolma [10], Nvidia's NeMo [11], DataComp's DCLM [12], and Unstructured.io [13]. While these projects have mainly focused on the preparation of data in creating GenAI models for natural languages (NLP), programming languages (Code), or both, recently the inverse problem has also been explored, namely, using LLMs in advanced data processing performed during down-stream AI applications like Fine-tuning, Instruction-tuning, and RAG [14] [15]. We will not consider these inverse projects in this paper. Before doing a direct feature-by-feature comparison of DPK with its closest and most relevant projects, namely, BigCode, DataTrove, and Dolma, as shown later in Table 1, we will do a descriptive comparison against a few other projects.

Nvidia Nemo-curator is focused on NLP only and leverages GPUs with DASK [16] for parallelization of Python and Nvidia's own RAPIDS GPU-accelerated libraries. The use of GPUs for jobs that can be done by CPUs may not be necessary. DPK achieves its high scalability via Ray and Spark frameworks and its automation via KFP.

DataComp-LM (DCLM) is a testbed for controlled dataset experiments to improve large language models. It allows for experimenting with a few data curation strategies such as deduplication, filtering, and data mixing at model scales, to achieve high-quality training datasets. DCLM focuses on NLP only and it uses Ray for scaling, but as compared to DPK, it has only a handful of transformation modules and does not target down-stream applications such as RAG and fine-tuning.

Unstructured.io, which is a full-featured "ingestion" engine for many file types used in LLM applications, is targeting RAG, therefore it includes chunking and embedding modules, but none of the other transforms that DPK has and has no real scaling for ingestion on a cluster (neither Ray nor Spark).

In Table 1, a comparison of DPK with its closest projects, across many dimensions, is shown. Note that DataTrove uses Slurm for scaling.

| Feature | BigCode | DataTrove | Dolma | DPK (Ours) |
|---|---|---|---|---|
| Target use case | Pretraining | Pretraining | Pretraining | Pretraining, Finetuning, RAG |
| Data modalities | Code | NLP | Both | Both |
| Scalability | Local /Spark | Local /Slurm | Local /Cluster | Local/Ray/Spark |
| Automation | CLI | CLI | CLI | CLI + KFP |
| User Personas | Data Scientist | Data Scientist | Data Scientist | Data Scientist/Application Developer |
| Easy to add new scalable modules | No | No | No | Yes |
| License | Apache 2.0 | Apache 2.0 | Apache 2.0 | Apache 2.0 |

Table 1 Comparing DPK with other similar Kits

## III. TOOLKIT DESIGN

DPK architecture is designed to enable developers to quickly create new transforms and easily deploy them to process data. It is designed to be data agnostic, and as such can support use cases in natural language and code data modalities. There are three primary blocks that build the architecture of DPK: Data Access, Transformation, and Runtime. We introduce them briefly here and then describe each of them in detail.

Data Access is a core element of the architecture and provides a general-purpose framework for identifying, selecting and reading, and writing data in a supported format. Checkpointing of performed work is supported independent of the Transform and Runtime components. Data Access is used by the Runtime and may be used by a Transform, for example to load a model from storage. Data access is configurable using the command line arguments. Its configuration is independent from the configuration of individual transforms and runtime.

Transformation components implement the specific operation on the data such as, data conversion, data deduplication, personally identifiable information identification, etc. Transforms are individually configurable using the command line arguments. For example, a resize transform will allow defining the size of the output data files. Transforms can also be executed in a sequence to form data preparation pipelines.

The Runtime identifies the execution environment for a given transform and starts one or more Transform Worker instances to operate on the data identified and provided by the Data Access component. With that, it distributes the work of transforming files to each of the Transform Workers, which operates on the assigned data.

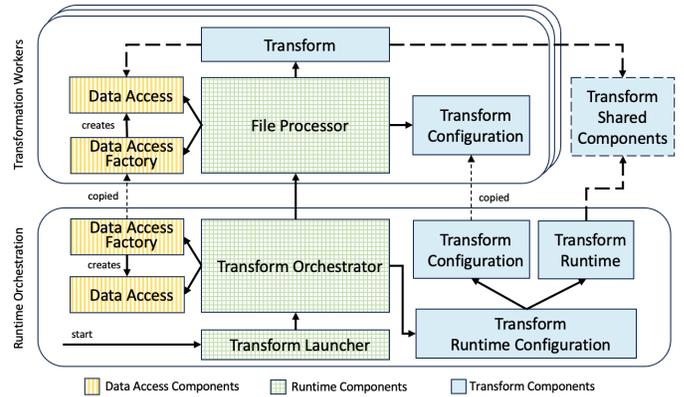

Fig. 2 The overall architecture of data preparation toolkit

### A. Data Access

Data used for processing can reside on a variety of storage devices, including local or distributed network file systems (NFS), S3-compatible storage, etc. To simplify usage of the multiple data storage mechanisms, the toolkit supports an abstraction layer, implemented using a Python class called DataAccess, that provides standardized data processing-oriented APIs, which are independent of the actual storage. The current implementation supports local file system and S3-compatible storage. Additional data access classes can be easily added to support user-specific storage types. Data Access components provide the ability to identify target data sources and read/write data from identified locations.

To enable runtime configuration, a factory for creating DataAccess instances (a Python class called DataAccessFactory), is used and configured using command line arguments. Factory configuration common across all DataAccess implementations includes:

- input path containing files to be processed
- output path to receive processed files
- file extensions to use as input

DataAccess configurations are customizable. For example, in the case of S3 data access, S3 credentials and endpoints are also configurable. The DataAccessFactory is shared across both the Runtime Orchestration and Transformation Workers components to enable the creation of the Data Access instance in each component. To add another DataAccess implementation, the DataAccessFactory class can be extended and used as an alternative in the Runtime.

Often restarts of large processing jobs are required. To support this, DataAccess supports checkpointing to determine which files in the input directory have not been processed to the output directory. This enables a checkpointed run to only process the unprocessed files saving perhaps large amounts of rework.

### B. Transformation

Data transformation motivates the need for a clean and concise yet powerful mechanism for manipulating arbitrary unstructured or structured data. It should support the following:

1:1 – a single data object is transformed into a single transformed data object. For example, annotating each row with a model score.

1: N – a single data object is transformed into multiple data objects. For example, splitting row data into multiple objects sized by row count.

N:1 – multiple data objects are aggregated into a single object. For example, joining row data into larger numbers of rows.

N: M – any number of data objects converted to any number of data objects. For example, sorting data into data objects of a specific type.

To meet all these needs, we define a base class, AbstractBinaryTransform, with the following methods:

`transform_binary(file_name, bytes_to_transform)` – The method transforms the given byte array into 0 or more byte arrays. The file name allows the transform to determine the format of the bytes (transforms may handle multiple data types). The method must return both a list of byte arrays and associated extensions to use when writing out the data, and optional metadata associated with the transformation that is collected by Statistics object (see below).

`flush_binary()` – supports the stateful accumulation of data across calls to the transform method. It returns the same data as the transform method. This can be useful, for example, to aggregate small parquet files into larger ones. In general, it should be called after all data has been processed through the transform method, as is done by the Runtime component.

Transform configuration is done at creation time via a dictionary of key/value pairs specific to the transform. The Runtime uses command line arguments to create the dictionary used to create and configure the transform.

To simplify working with Arrow tables (typically read from .parquet files), the toolkit includes an AbstractTableTransform class to enable the transformation of Arrow tables. `transform(table, file_name)` and `flush()` methods are available with semantics similar to the binary versions.

Additional transform classes can be created (similar to AbstractTableTransform) to simplify the processing of the specific file types.

Command line-based configuration and runtime enablement of a transform is provided through a Transform's Configuration. This is implemented via a base Python class called TransformConfiguration. This class defines:

- The name of the transform as reported by the Runtime
- The class name of the Transform implementation
- Methods to define and validate the command line transform configuration arguments used to instantiate the Transform.

The functionality of this class is separated from the transform class itself to avoid premature instantiating the transform. Because the transform implementation might be unpicklable and may have large initialization costs such as model loading or other expensive configuration requirements on either Memory or CPU, we only want to create the instance of the transform where it will be used.

### C. Runtime

Runtimes are responsible for establishing one or more transform operating environments, assigning work to them and monitoring and reporting progress. DPK currently includes three different runtime implementations:

- Pure Python: runs transforms within the Python process. Additionally, there is support for Python multiprocessing.
- Ray: runs transforms in Ray Actors either using local or remote Ray cluster.
- Spark: runs transforms using either local or remote Spark cluster.

These runtimes allow the user the flexibility to deploy the computation across compute facilities ranging from local single computers to Kubernetes clusters consisting of thousands of nodes. To simplify testing on Kubernetes clusters, DPK provides instructions and scripts for deploying itself on a simple Kind cluster. This can be used for testing execution on Kubernetes locally.

Additionally, each runtime allows transforms to leverage specific functionality that it can provide, for example, shared classes for Python processes, shared actors and object store for ray, etc. To control these functions each transform can implement runtime support, specific for a given runtime, for example *DefaultPythonTransformRuntime* for Python, *DefaultRayTransformRuntime* for Ray, and *DefaultSparkTransformRuntime* for Spark. Transform-specific implementation of these classes allows transform implementers to use such native runtime facilities and pass them as parameters to transform execution. This capability is used by some of the transforms, for example Doc ID, dedup, etc to keep running execution state.

There are three primary components provided by each of the runtimes – launcher, orchestrator, and workers.

The Transform Launcher is the entry point to running a transform and is usually started from a main() initiated from the command line. It accepts a set of command line arguments to configure the Runtime, Data Access Factory, and Transform itself.

After the configuration of the components by the Transform Launcher, the launcher optionally initializes runtime (for example creates a local Ray/Spark cluster or connects to an existing one) and calls the runtime-specific Transform Orchestrator.

The Transform Orchestrator uses the Transform Runtime to establish the optional Transform Shared Components (using runtime support classes, which are part of transform implementation) and uses the Data Access to identify the set of input files that need to be processed according to configuration, including location, file extensions, and checkpointing. It then creates a shared statistics object (actor) and workers – in the case of Python, this optionally creates a multiprocessing pool [17], in the case of Ray it creates worker Actors, in the case of Spark, it creates executors. The Transform Orchestrator receives copies of the Data Access Factory, Transform Configuration and creates the File Processor(s).

The File Processor instantiates the Transform and Data Access after which it begins receiving names of files to process from the Transform Orchestrator, generally one at a time. Upon receipt of a file name, the File Processor reads the file and provides the contents to the Transform for processing. It then writes execution results back to storage using the Data Access APIs and adds processing metadata to the shared statistics.

After exhausting the list of files to process, the Transform Orchestrator indicates completion to the File Processor, which then invokes its Transform flush method and writes the results to storage. Finally, the Transform Orchestrator writes accumulated metadata from shared Statistics to Data Access storage and the process is complete. The DPK architecture provides significant flexibility and power to deploy computation using a well-defined architecture consisting of data access, computation/transformation, and deployment and management of these components. DPK supports a range of computing infrastructures including pure Python, Ray, and Spark, and provides scalability (local to large Kubernetes clusters). With this flexibility and configurability, DPK can be applied to a wide range of use cases in LLM data preparation ranging from full pre-training to RAG and fine-tuning.

## IV. AVAILABLE TRANSFORMS

The toolkit comes with several prebuilt transforms that can be divided into four main categories: (A) Data Ingestion, (B) Universal, (C) Code, and (D) Language. In addition to pure Python implementations, for most transforms, we also provide Ray-based KFP for automation. For a few transforms, a Spark-based version has also been implemented as a reference for the community to add more. Table 2 shows the current list of transforms in DPK.

## V. AUTOMATION

To offer a highly scalable and easily executable solution with a history of previous executions, all DPK transformations can be executed using KFP. KFP is a popular platform for building and deploying scalable machine learning (ML) workflows on Kubernetes. It allows for defining, orchestrating, and managing end-to-end ML workflows in a cloud-native way, ensuring that data processing tasks can be efficiently executed, monitored, and scaled.

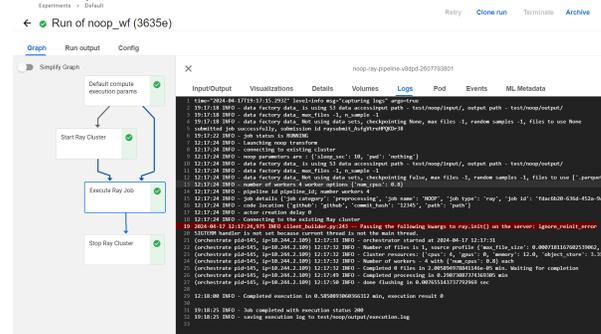

Fig. 3 Simple pipeline execution

The integration with KFP offers several benefits:
- **Scalability:** KFP runs on Kubernetes, so it can scale to handle large datasets and complex workflows. In the EXPERIMENTAL EVALUATION section, we describe how this solution was scaled to hundreds of nodes with thousands of CPU cores and tens of terabytes of RAM.
- **Modularity**: Our workflows are broken down into reusable shared components, making it easy to build and debug complex pipelines by chaining together individual tasks. For all transforms we use the same shared components: "Compute execution parameters", "Start a Ray cluster", "Execute a Ray job" and finally "Stop/destroy the Ray cluster. The last step is executed regardless of whether the Ray job execution succeeds or fails. You can see the execution status of a successful job and its logs in Fig. 3.
- **History and Reproducibility**: By maintaining a history of executions, KFP ensures the reproducibility of experiments. Our KFP deployments are integrated with standard logging systems such as Prometheus, Thanos, and Grafana. However, to guarantee independence on these systems, all execution logs and metrics are automatically stored in an S3 object store.
- **Visualization**: It offers a UI for monitoring pipeline runs, visualizing results, and troubleshooting issues.

As described above, DPK executes each transform in a "simple" KubeFlow pipeline. For the cases when we need to execute several transforms, DPK introduces the concept of a "super" pipeline, which combines several of these simple (nested) pipelines. Fig. 4 demonstrates the "super" pipeline for Code preprocessing, where each "super" pipeline step is a nested "simple" pipeline. Both the pipelines and the super pipelines provide a no-code interface to run the data preparation pipelines in production settings.

| Modules | Python-only | Ray | KFP on Ray | Spark | Short Description/Referenced Work |
|---|---|---|---|---|---|
| **Data Ingestion** | | | | | |
| Code (from zip) to Parquet | ✓ | ✓ | ✓ | | Converts ZIP files containing programming files (.py, .c, .java, etc), into Parquet format |
| PDF to Parquet | ✓ | ✓ | ✓ | | Based on Docling Library [18] |
| HTML to Parquet | ✓ | | | | Based on Trafilatura [19] |
| **Universal (Code & Language)** | | | | | |
| Exact Dedup Filter | ✓ | ✓ | ✓ | | Identifies (and removes) duplicate records from the sets of data, based on a hash-based matching of documents [1] |
| Fuzzy Dedup Filter | | ✓ | ✓ | | Based on the MinHash and LSH algorithm [1] |
| Unique ID Annotation | ✓ | ✓ | ✓ | ✓ | Adds a document identification (unique integers and content hashes |
| Filter on Annotations | ✓ | ✓ | ✓ | ✓ | Provides SQL-based filtering of records, to remove documents that do not conform to the user criteria |
| Profiler | | ✓ | ✓ | | Allows the profiling of the source data for evaluating its properties |
| Resize | ✓ | ✓ | ✓ | | Allows the alignment of file sizes (both splitting larger ones and combining smaller ones) |
| Tokenizer | ✓ | ✓ | ✓ | | Based on the HuggingFace Tokenizer [1] |
| No-op/template | ✓ | ✓ | ✓ | ✓ | Simple 1:1 transform, as a test case for development |
| **Language-only** | | | | | |
| Language Identification | ✓ | ✓ | ✓ | | Identifies the language of each text with a confidence score, using the fasttext language identification model [20] |
| Document Quality | ✓ | ✓ | ✓ | | Calculates and annotates several metrics related to the document, which are useful to determine the quality of the document [21] |
| Document Chunking for RAG | ✓ | ✓ | ✓ | | Chunks documents by leveraging HierarchicalChunker in [22]. It also relies on documents converted with the Docling library [18]. |
| Text Encoder | ✓ | ✓ | ✓ | | Uses sentence encoder models to create embedding vectors of the text in each row of the input Parquet table. These vectors are useful for tasks which are at the core of Retrieval-Augmented Generation (RAG) applications. |
| PII Annotator/Redactor | ✓ | ✓ | ✓ | | Redacts Personally Identifiable Information (PII) from the input data. The transform leverages the Microsoft Presidio SDK [23][24] for PII detection and uses the Flair recognizer for entity recognition [25] |
| **Code-only** | | | | | |
| Programming Language Annotation | ✓ | ✓ | ✓ | | Allows the user to specify the programming languages for which the data should be identified |
| Code Quality Annotation | ✓ | ✓ | ✓ | | Captures code-specific metrics of input data. The implementation is borrowed from the work done in StarCoder [26] |
| Malware Annotation | ✓ | ✓ | ✓ | | Scans the 'contents' column of an input table using ClamAV [27], and outputs corresponding tables containing 'virus_detection' column |
| Header Cleanser | ✓ | ✓ | ✓ | | Detects and removes license and copyright information from code files. It leverages the ScanCode Toolkit [28] |
| Semantic File Ordering | | ✓ | | | Sorts the repo content by file path or in semantic ordering. For more information on this transform, refer to [29] |

*Table 2 DPK transforms*

## VI. HOW TO BRING YOUR OWN TRANSFORM

DPK is intended to be an extensible library allowing the creation of custom transforms that can then be applied to data using one of the runtimes. Here we provide the classic "hello world" example to illustrate some of the steps involved in writing a new transform. Although the Data Prep Kit supports the transformation of arbitrary byte arrays, we will focus here on the specialization for transforming PyArrow Table objects, typically read from parquet files.

As a reminder, the new transform will add a column containing a hello message defined using command line arguments. This new transform is implemented by providing the two classes HelloTransform and HelloTransformConfiguration.

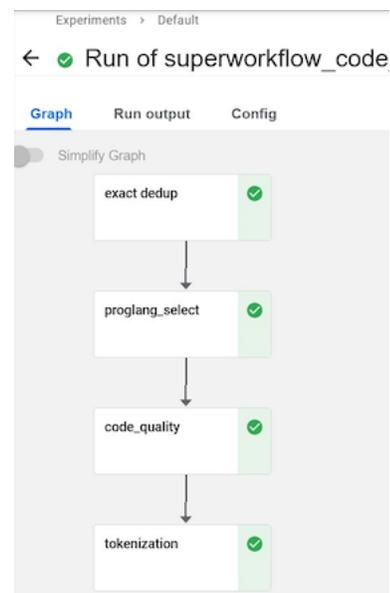

*Fig. 4 Super pipeline*

HelloTransform extends the core AbstractTableTransform to enable configuration through an initializer and provides a transform() method implementation that adds the new column to the provided PyArrow Table object.

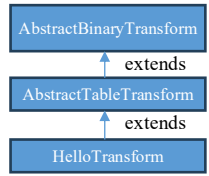

All transforms receive a dictionary of configuration data in which the keys a strings and values are arbitrary data that is specific to the transform implementation. In our case, we will allow the transform to be configured with the name of the entity to be greeted. The class declaration and initializer are as follows:

```python
class HelloTransform(AbstractTableTransform):
    def __init__(self, config:dict):
        self.who = config.get("who", "World")
        self.column_name = config.get("column_name", "greeting")
```

The initializer allows two configurations – the name to be greeted and the name of the new column in the Table to hold the greetings. Next, we implement the transform() method to create the column of greetings and add it to the table.

The transform() method takes the in-memory PyArrow Table and the file name from which the table was read, typically a parquet file. Most transforms will not make use of the file name, but it is provided for advanced usage.

```python
def transform(self, table: pa.Table, file_name: str = None) -> (
        tuple)[list[pa.Table], dict[str, Any]]:
    new_column = ["Hello " + self.who + "!"] * table.num_rows
    table = TransformUtils.add_column(table=table, name=self.column_name,
                                      content=new_column)
    return [table], {"nrows": table.num_rows}
```

The code above creates the greeting using the configured entity name (i.e. self.who) and creates a list of greetings of length equal to the number of rows in the given Table. We then use a DPK convenience method to create a new table from the input and the new column of greetings. Finally, we return the table and optional metadata to be associated with each call to transform().

To support transforms that need to transform one table into multiple tables (e.g. resize with may split tables into multiple), the transform() method returns a list of Table objects. When a runtime receives more than one table, it appends an index to the output file names.

Metadata returned by a transform is aggregated by the runtimes and included in the metadata.json written by the runtime at the end of a transform job. In the example above, we simply include the number of rows which will then be accumulated across all Tables processed.

To enable a transform to be run in a DPK runtime, a transform configuration class, and a runtime configuration class are needed. The Runtime Configuration allows runtime-specific configurations and holds the Transform Configuration for which the runtime is configured to run.

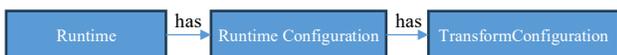

A transform's configuration class extends the TransformConfiguration class to define the associated transform implementation (i.e., HelloTransform), command line options to configure that transform, and finally the name of the transform as reported by the runtime. The transform configuration class for HelloTransform begins as follows to define the name of the as "hello" and the associated transform class as HelloTransform.

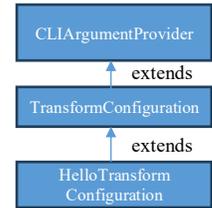

```python
class HelloTransformConfiguration(TransformConfiguration):
    def __init__(self):
        super().__init__(
            name="hello",
            transform_class=HelloTransform,
        )
```

To define the command line arguments that configure the transform, the CLIArgumentProvider methods add_input_params() and apply_input_params() must be defined. Below we add the configuration for both who will be greeted and the name of the column to hold the greetings.

```python
def add_input_params(self, parser: argparse.ArgumentParser) -> None:
    parser.add_argument(
        *name_or_flags: "--ac_who",
        type=str, required=False, default="World",
        help="Who to say hello to."
    )
    parser.add_argument(
        *name_or_flags: "--ac_column_name",
        type=str, required=False, default="greeting",
        help="Name of column to add"
    )

def apply_input_params(self, args: argparse.Namespace) -> bool:
    dargs = vars(args)
    self.params = {
        "who": dargs.get("ac_who", None),
        "column_name": dargs.get("ac_column_name", None)
    }
    return True
```

Next, we define the transform's Python runtime configuration class which holds the HelloTransformConfiguration class and a main() to trigger the running of the transform on the input data according to the command line options.

```python
class HelloPythonConfiguration(PythonTransformRuntimeConfiguration):
    def __init__(self):
        super().__init__(transform_config=HelloTransformConfiguration())

if __name__ == "__main__":
    launcher = PythonTransformLauncher(HelloPythonConfiguration())
    launcher.launch()
```

To run in a Ray runtime, for example, replace "Python" with "Ray" above. Finally, run the transform on parquet files in a local directory called input and place the transformed data in a directory called output:

```
% Python hello_transform.py  --data_local_config \
    "{ 'input_folder': 'input', 'output_folder': 'output'}" \
    --who Universe  --column_name hello
```

## VII. EXPERIMENTAL EVALUATION

Scalability is a critical determinant for the suitability of transforms for real-world applications where large datasets and high-performance computing are common. In this section, we present evidence supporting the scalability and resource efficiency of DPK.

To this end, we conducted a series of experiments evaluating the performance of individual transforms under various conditions. These experiments encompassed running transforms independently on smaller datasets with limited computational infrastructure and on large-scale clusters with numerous CPU cores. The results indicate that DPK maintains its high level of performance consistently across these diverse settings.

To investigate the impact of logic complexity on the performance of transforms, we examine the average throughput of selected transforms executed on a single node. This experiment utilizes a node with 16 CPU cores and 64GB of RAM, allowing us to explore the relationship between complexity and execution in a controlled low-resource environment.

The average throughput of the transforms running on a single node expressed in megabytes per second is depicted in Fig. 5. We observe that the intricacy of the transform logic has a direct influence on the time required by each transform. There is a default "sleep time" for the noop transform that has resulted in its less-than-expected throughput.

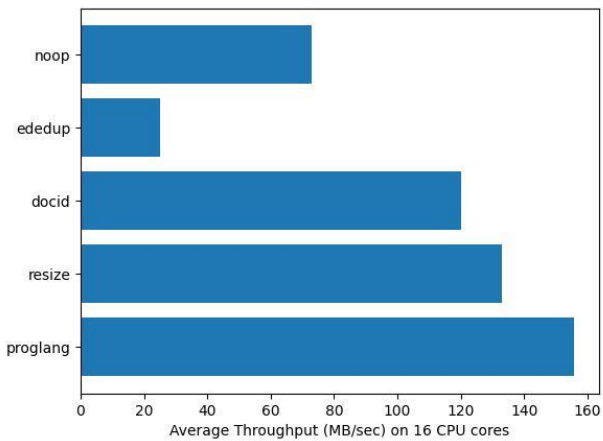

*Fig. 5 Average throughput recorded by Transforms (single-node)*

Having established the performance of transforms on a single node, we next evaluate the scalability of DPK in a cluster setting. To do this, we conducted a series of small to medium-scale experiments on a cluster of 40 nodes each equipped with 48 CPU cores, and 384GB of RAM. The experiments were designed to test the transforms' performance on varying dataset sizes and increasing node counts. The dataset used for the experiments consists of common crawl snapshots.

To evaluate the scalability of the transforms, we examine the impact of three categories of transforms on data processing in DPK. The first category (C1) consists of a basic transform: *doc ID*. Doc ID annotates existing rows with a unique integer without any data or schema transformation. The second category (C2) includes two transforms that perform file manipulation beyond annotation: *resize* and *ededup*. Resize splits or merges files into a target size using buffer manipulation, while *ededup* finds and eliminates duplicates in an input data corpus. Finally, *language identification* stands alone in the third category (C3), inferring a model for identification.

We evaluate the throughput of the transforms, expressed in units of terabytes per minute as shown in Fig. 6. We further assert that there is a direct relationship between the effectiveness of the transforms in handling increasing volumes of data and the complexity of the transform logic. For instance, the *lang ID* transform from C3, which inferences the FastText language identification model [20], can lead to lower throughput compared to transforms in lower categories such as the *Doc ID* transform.

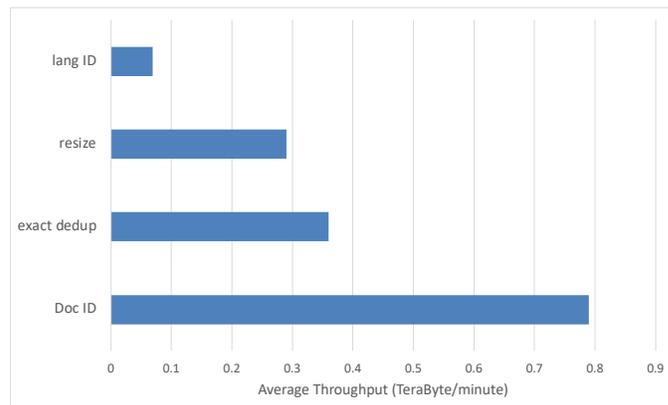

*Fig. 6 Average throughput recorded by Transforms (cluster)*

Similarly, we measure the time required to process a fixed amount of data regardless of the overall processing rate, measured in "Execution Time per TeraByte" as shown in Fig. 7. The scalability of DPK is evident in the proportional reduction of execution time for transforms as the number of processing nodes is increased, demonstrating a strong ability to handle distributed workloads, towards the end the performance plateaus which is because the dataset no more stays feasibly sized for that number of CPUs. With an average of ~40% (accounting for I/O and network variations) reduction in Execution time per TeraByte of data as the number of CPU cores doubles, we show that DPK transforms perform well in low-resource environments and also scale well with increased resource availability.

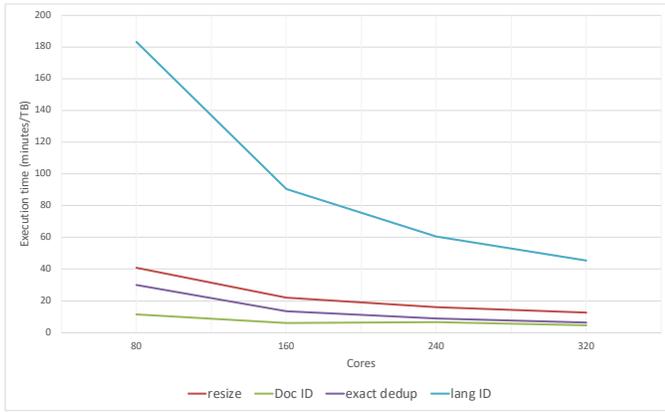

*Fig. 7 Execution Time per TeraByte of Data vs. Total Number of CPU cores for individual transforms*

The results also indicate that I/O bound transforms (C1) have the least impact on scalability, while demonstrating moderate influence on complex file manipulation transforms (C2), and the most substantial impact on scaling for transforms requiring the most computational resources to infer a model (C3).

The findings highlight the importance of considering the impact of transform logic on scaling when using DPK for data processing. By carefully selecting the appropriate transforms and optimizing their use, users can achieve more efficient and effective data processing results.

| Transform Name | Input Data Size | Percentage of data filtered | Compute Time (minutes) | Number of CPU Cores |
|---|---|---|---|---|
| E-Dedup | 2TB | 16.14 | 38.15 | 480 |
| F-Dedup | 2TB | 24.3 | 1511.65 | 480 |
| E-Dedup | 332GB | 3.3 | 5.2 | 320 |
| F-Dedup | 332GB | 4.9 | 107.49 | 320 |

*Table 3 Exact and Fuzzy Dedup Data Processing Metrics: F-Dedup configuration: {"num_permutations": 64, "threshold": 0.8, "shingles_size": 5}*

Finally, we examine the practical application of two transforms that facilitate data removal in DPK. This is an important consideration in many applications, as the ability to remove redundant data can significantly improve data quality. The results presented in Table 3 demonstrate that DPK is robust and adaptable, exhibiting consistent performance regardless of the size of the input data or resources available. This consistency is a testament to DPK's ability to effectively handle diverse processing requirements, making it a valuable tool for various applications.

## VIII. CONCLUSION

This paper presents Data Preparation Kit for LLM applications. DPK is flexible (runs on different platforms), and extensible to add new scalable modules without having deep Ray and/or Spark expertise. It also provides out-of-the-box automation for existing as well as newly added modules. DPK is a very useful toolkit for all the users who aim to prepare data but would like to use already implemented transforms and at the same time be able to easily customize or extend the toolkit to fit their needs.

DPK comes with modules that users can either leverage independently or use in a pipelined fashion. Automation is designed into the DPK which enables users to easily scale up their workload on clusters through the KFP dashboard without writing any code or worrying about configuring and maintaining any cluster. Moreover, the same automation is inherited by any new modules that the user would like to add to the toolkit. Modules from DPK have been used with automation at scale in preparing data for IBM Granite Models thus we strongly feel that it will prove to be very valuable to the larger LLM data engineering community. Our future plans include the expansion of the DPK capabilities with support for new data modalities, additional scalable runtimes, and new readily usable transforms.

## REFERENCES

[1] Mayank Mishra et al, "Granite Code Models: A Family of Open Foundation Models for Code Intelligence," in *Arxiv*, 2024.

[2] "IBM Granite Models," 2024. [Online]. Available: https://huggingface.co/ibm-granite.

[3] Guilherme Penedo et al, "The RefinedWeb Dataset for Falcon LLM: Outperforming Curated Corpora with Web Data, and Web Data Only," in *Arxiv*, 2023.

[4] Hima Patel et al, "A Data-centric AI Framework for Automating Exploratory Data Analysis and Data Quality Tasks," *ACM Journal of Data and Information Quality,* pp. 1-26, 2023.

[5] Moritz et al, "Ray: A distributed framework for emerging {AI} applications.," in *13th USENIX symposium on operating systems design and implementation*, 2018.

[6] Matei Zaharia et al, "Apache spark: a unified engine for big data processing.," in *Communications of the ACM*.

[7] E. Bisong, Kubeflow and kubeflow pipelines." Building Machine Learning and Deep Learning Models on Google Cloud Platform: A Comprehensive Guide for Beginners, Springer, 2019.

[8] BigCode, 2023. [Online]. Available: https://www.bigcode-project.org/.

[9] Guilherme Penedo et al, "Datatrove: large scale data processing,," 2024.

[10] Soldaini et al, "Dolma: An open corpus of three trillion tokens for language model pretraining research.," in *Arxiv*, 2024.

[11] G. B. Harper et al, "NeMo: a toolkit for Conversational AI and Large Language Models," 2024. [Online]. Available: https://github.com/NVIDIA/NeMo.

[12] Li et al, "DataComp-LM: In search of the next generation of training sets for language models," 2024.

[13] "Get your data RAG ready," 2024. [Online]. Available: https://unstructured.io/.

[14] M. Vangeli, "Large Language Models as Advanced Data Preprocessors: Transforming Unstructured Text into Fine-Tuning Datasets," 2024.

[15] Haochen Zhang et al, "Jellyfish: A Large Language Model for Data Preprocessing," 2024.

[16] M. Rocklin, "Dask: Parallel Computation with Blocked algorithms and Task Scheduling," in *PROC. OF THE 14th PYTHON IN SCIENCE CONF. (SCIPY 2015)*, 2015.

[17] "Python Multiprocessing Pool," [Online]. Available: https://superfastPython.com/multiprocessing-pool-Python/.

[18] Auer et al, "Docling Technical Report," in *Arxiv*, 2024.

[19] A. Barbaresi, "Trafilatura: A Web Scraping Library and Command-Line Tool for Text Discovery and Extraction," in *Proceedings of ACL/IJCNLP 2021: System Demonstrations*, 2021.

[20] L. Mouselimis, "Library for efficient text classification and representation learning," 2022. [Online]. Available: https://fasttext.cc/.

[21] Jack W. Rae, "Scaling Language Models: Methods, Analysis & Insights from Training Gopher," in *Arxiv*, 2021.

[22] "Quackling," 2024. [Online]. Available: https://github.com/DS4SD/quackling.

[23] "Context aware, pluggable and customizable data protection and de-identification SDK for text and images," [Online]. Available: https://github.com/microsoft/presidio/.

[24] "Presidio," [Online]. Available: https://github.com/microsoft/presidio/.

[25] Alan Akbik et al, "FLAIR: An Easy-to-Use Framework for State-of-the-Art NLP," in *Proceedings of the 2019 Conference of the North American Chapter of the Association for Computational Linguistics (Demonstrations)*, 2019.

[26] Starcoder: Li, Raymond, Loubna Ben Allal, Yangtian Zi, Niklas Muennighoff, Denis Kocetkov, Chenghao Mou, Marc Marone et al. "Starcoder: may the source be with you!." *arXiv preprint arXiv:2305.06161* (2023)

[27] "ClamAV," [Online]. Available: https://www.clamav.net.

[28] "ScanCode," [Online]. Available: https://scancode-toolkit.readthedocs.io/en/stable/.

[29] Matt Stallone et al, "Scaling Granite Code Models to 128K Context," in *Arxiv*, 2024.